\documentclass[lettersize,journal]{IEEEtran}
\usepackage{algorithmic}
\usepackage{algorithm}
\usepackage{array}
\usepackage{textcomp}
\usepackage{stfloats}
\usepackage{url}
\usepackage{verbatim}
\usepackage{graphicx}
\usepackage{cite}
\usepackage{amsmath,amssymb,amsfonts}
\usepackage{subfigure}
\usepackage{xcolor}
\usepackage{stmaryrd} 
\newcommand{\argmin}{\operatornamewithlimits{\arg\,\min}}
\newcommand{\lb}{\llbracket}
\newcommand{\rb}{\rrbracket}

\begin{document}

\title{When Evolutionary Computation Meets Privacy}

\author{Bowen Zhao,~\IEEEmembership{Member,~IEEE,},
        Wei-Neng Chen,~\IEEEmembership{Senior Member,~IEEE},
        Xiaoguo Li,
        Ximeng Liu,~\IEEEmembership{Senior Member,~IEEE},
        Qingqi Pei,~\IEEEmembership{Senior Member,~IEEE},
        and Jun Zhang,~\IEEEmembership{Fellow,~IEEE},
\thanks{This research is partly supported by the National Key Research and Development Program of China (No. 2022YFB3102700) and the National Natural Science Foundation of China (Nos. 62202358, 61976093, and 62072109).}
\thanks{Bowen Zhao is with Guangzhou Institute of Technology, Xidian University, Guangzhou, China. E-mail: bwinzhao@gmail.com}
\thanks{Wei-Neng Chen is with the School of Computer Science and Technology, South China University of Technology, Guangzhou, China. E-mail: cwnraul634@aliyun.com}
\thanks{Xiaoguo Li is with the School of Computing and Information Systems, Singapore Management University, Singapore, Singapore. E-mail: xiaoguoli@smu.edu.sg}
\thanks{Ximeng Liu is with the College of Computer and Data Science, Fuzhou University, Fujian, China, and Cyberspace Security Research Center, Peng Cheng Laboratory, Shenzhen. snbnix@gmail.com}
\thanks{Qingqi Pei is School of Telecommunications Engineering, Xidian University, Xian, China. E-mail: qqpei@mail.xidian.edu.cn}
\thanks{Jun Zhang is with Zhejiang Normal University, Jinhua, China, and
also with Hanyang University, Ansan, South Korea.}
\thanks{Corresponding authors: Wei-Neng Chen, Bowen Zhao}
\thanks{Manuscript received XX XX, 202X; revised XX XX, 202X.}}

\markboth{Journal of \LaTeX\ Class Files,~Vol.~XX, No.~XX, August~202X}%
{Shell \MakeLowercase{\textit{et al.}}: A Sample paper Using IEEEtran.cls for IEEE Journals}


\maketitle

\begin{abstract}
Recently, evolutionary computation (EC) has been promoted by machine learning, distributed computing, and big data technologies, resulting in new research directions of EC like distributed EC and surrogate-assisted EC. These advances have significantly improved the performance and the application scope of EC, but also trigger privacy leakages, such as the leakage of optimal results and surrogate model.
Accordingly, evolutionary computation combined with privacy protection is becoming an emerging topic. 
However, privacy concerns in evolutionary computation lack a systematic exploration, especially for the object, motivation, position, and method of privacy protection.
To this end, in this paper, we discuss three typical optimization paradigms  (i.e., \textit{centralized optimization, distributed optimization, and data-driven optimization}) to characterize optimization modes of evolutionary computation and propose BOOM (i.e., \textit{o\textbf{B}ject, m\textbf{O}tivation, p\textbf{O}sition, and \textbf{M}ethod}) to sort out privacy concerns in evolutionary computation. Specifically, the centralized optimization paradigm allows clients to outsource optimization problems to the centralized server and obtain optimization solutions from the server. While the distributed optimization paradigm exploits the storage and computational power of distributed devices to solve optimization problems. Also, the data-driven optimization paradigm utilizes data collected in history to tackle optimization problems lacking explicit objective functions. Particularly, this paper adopts BOOM to characterize the object and motivation of privacy protection in three typical optimization paradigms and discusses potential privacy-preserving technologies balancing optimization performance and privacy guarantees in three typical optimization paradigms. Furthermore, this paper attempts to foresee some new research directions of privacy-preserving evolutionary computation.
\end{abstract}


\section{Introduction}
\IEEEPARstart{E}{volutionary} computation is a batch of optimization algorithms inspired by biological evolution to solve real-world optimization problems such as science, engineering, and technology \cite{bujok2021differential,kudela2022critical}. Technically, given an optimization problem, evolutionary computation initializes a set of candidate solutions via a population and iteratively updates them to approximate the optimal solution. Then, the population gradually evolves to improve the fitness of the candidate solution through evolutionary operations, such as crossover, mutation, fitness evaluation, and selection. Recently, evolutionary computation has been extensively used in routing planning, industrial design, tuning hyperparameters, resource scheduling, training neural networks, etc \cite{vargas2022review}.

On the one hand, the rapid development of distributed and parallel computing paradigms like supercomputing, cloud computing, edge computing, etc., extends the application scope of evolutionary computation as well as improves the performance of evolutionary computation. 
Here we present three typical optimization paradigms. The first paradigm is to use computing platforms such as cloud computing to provide complex optimization services based on evolutionary computing. The cloud server is generally responsible for maintaining a population and its candidate solutions and approximating the optimal solution \cite{zhang2016self,jiang2020privacy}.
The second paradigm is to use a group of distributed agents to participate in the evolutionary optimization process of evolutionary computation. 
A cluster of distributed agents with storage and computational power forms a population and jointly generates candidate solutions and approximates the optimal solution \cite{gong2015distributed,yang2020distributed}.
The third paradigm is that evolutionary computation is used to deal with complex data-driven optimization problems, i.e., data-driven evolutionary optimization, but data is collected and stored in a distributed manner in a network environment. The data-driven optimization paradigm takes advantage of historical data collected in simulations, physical experiments, production processes, or daily life to train a surrogate model and evaluate the fitness of candidate solutions \cite{jin2019data,wei2021classifier}.

On the other hand, three typical optimization paradigms trigger privacy concerns. The centralized cloud server is usually untrustworthy and frequently leaks user data, such as Microsoft Azure and Amazon web services breaching user data \cite{zuo2019does}. In the centralized optimization paradigm \cite{zhang2016self,jiang2020privacy}, the centralized server is extremely likely to leak the optimization result. The distributed optimization paradigm requires multiple participants/agents/devices to approximate locally optimal solutions and generates a globally optimal solution based on those locally optimal solutions \cite{gong2015distributed,yang2020distributed}. Also, each participant/agent/device needs to obtain the globally optimal solution and then update her local solution. In this case, the locally optimal solution and the globally optimal solution may be leaked. The data-driven optimization paradigm evaluates the fitness through a surrogate model trained by historical data \cite{jin2019data,wei2021classifier}, thus, it might reveal the surrogate model or the inference data (a possible solution for an optimization problem).

The combination between evolutionary computation and the technology of privacy protection provides a potential solution and raises attention. Either machine learning or evolutionary computation is seen as the subfield of artificial intelligence. To tackle privacy concerns in machine learning, privacy-preserving machine learning integrating privacy-preserving technology into machine learning has been extensively explored \cite{bonawitz2019towards,cabrero2021sok}. Encryption is a direct but effective privacy-preserving technology for protecting private data, which enables access control \cite{zhang2021attribute}, data confidentiality guarantee \cite{boulemtafes2020review}, and even computation over encrypted data directly \cite{li2019privacy}. Inspired by privacy-preserving machine learning, several schemes combine privacy-preserving technology and evolutionary computation \cite{jiang2020privacy,zhao2022primpso,liu2022secure} to address the challenge of privacy leakages in evolutionary computation.

In contrast to privacy-preserving machine learning, there are several reasons that hinder the full exploration of privacy-preserving evolutionary computation. \textbf{Fistly}, the object of privacy protection in evolutionary computation is not clear. It is unclear whether the object of privacy protection in evolutionary computation is the optimization problem, candidate solutions, the optimal solution, or an objective function. \textbf{Secondly}, as the object of privacy protection is unclear, it first is unaware of what data to protect, and then it lacks the motivation to protect private or sensitive data. \textbf{Thirdly}, the position using privacy-preserving technology is not transparent. Given an optimization problem, it is not transparent whether privacy-preserving techniques are adopted in clients holding optimization problems, centralized servers, distributed devices, or entities holding historical data. \textbf{Lastly}, it is significantly challenging to balance the performance of evolutionary computation and privacy protection. Privacy-preserving evolutionary computation based on existing privacy-preserving technology may be limited by the performance and type of computation as existing privacy-preserving technology either suffers from a high computation burden or supports the limited types of computation on encrypted data.

In this paper, we try to systematically sort out privacy concerns in the evolutionary computation domain to bridge the gap between evolutionary computation and privacy protection. Specifically, we first formalize three typical optimization paradigms of evolutionary computation. Then, for the first time, we present BOOM \footnote{BOOM: o\underline{\textbf{B}}jective, m\underline{\textbf{O}}tivation, p\underline{\textbf{O}}sition, and \underline{\textbf{M}}ethod}, a general framework to characterize privacy concerns in evolutionary computations and verify it in three typical optimization paradigms. The technical contributions of this paper are three-fold.
\begin{itemize}
    \item We introduce three typical optimization paradigms of evolutionary computation inspired by evolutionary computation modes, i.e., centralized optimization, distributed optimization, and data-driven optimization. 
    \item We propose a general framework BOOM to characterize the research problems in three typical optimization paradigms of privacy-preserving evolutionary computation, including the object, motivation, position, and method of privacy protection in the evolution computation domain for the first time.
    \item We identify several meaningful but challenging research directions in the privacy-preserving evolutionary computation domain.
\end{itemize}

The rest of this paper is organized as follows. In Section II, we formalize three typical optimization paradigms of evolutionary computation. In Section III, we define BOOM, a general framework to cover privacy concerns in evolutionary computation. We elaborate on how to use the proposed BOOM to identify privacy concerns in a centralized optimization paradigm, a distributed optimization paradigm, and a data-driven optimization paradigm in Section IV, Section V, and Section VI, respectively. In section VII, we foresee future research directions of privacy protection in evolutionary computation. Finally, we conclude the paper in Section VIII.

\section{Typical Paradigm of Evolutionary Computation}
\begin{figure}[t]
    \centering
    \subfigure[Centralized optimization paradigm]{
    \includegraphics[width=0.83\linewidth]{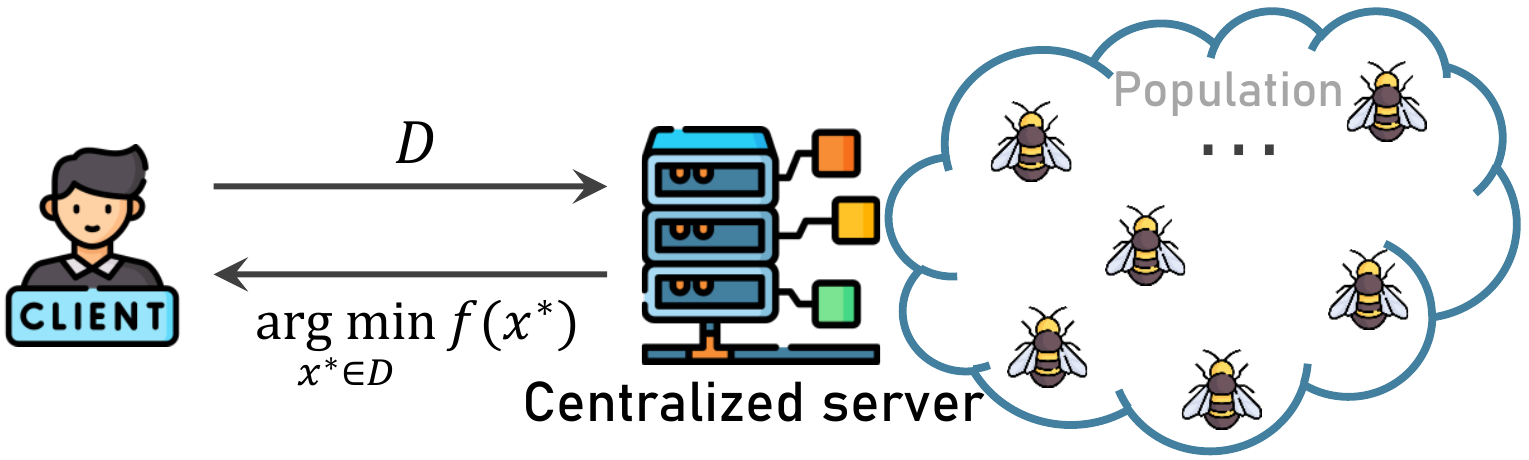}}
    \hspace{0.225in}
    \subfigure[Distributed optimization paradigm]{
    \includegraphics[width=0.83\linewidth]{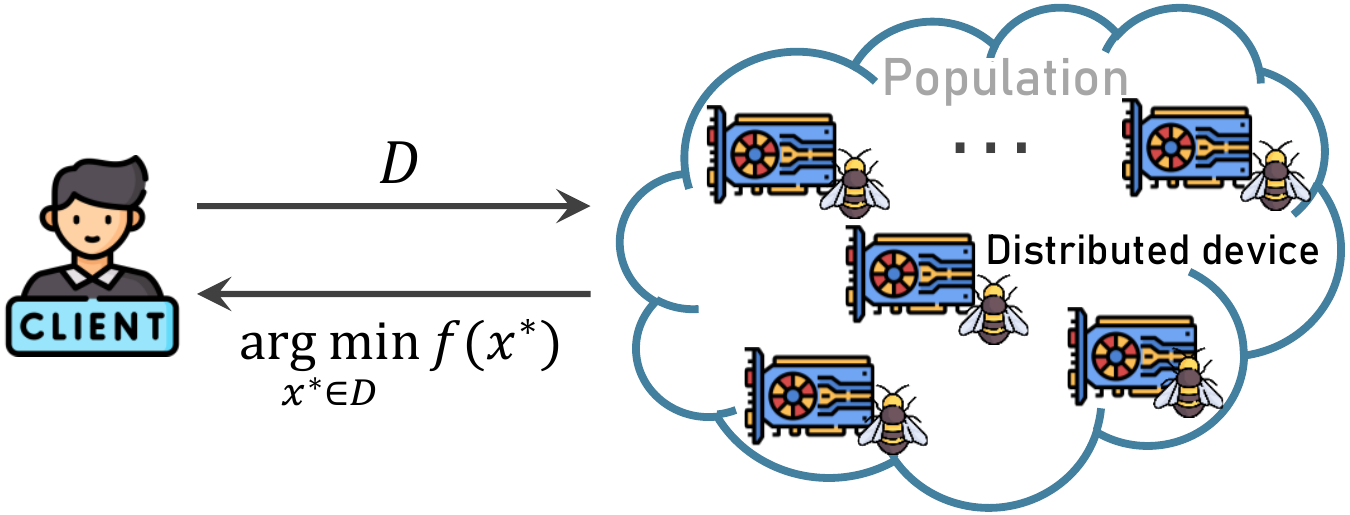}}
    \hspace{0.225in}
    \subfigure[Data-driven optimization paradigm]{
    \includegraphics[width=0.83\linewidth]{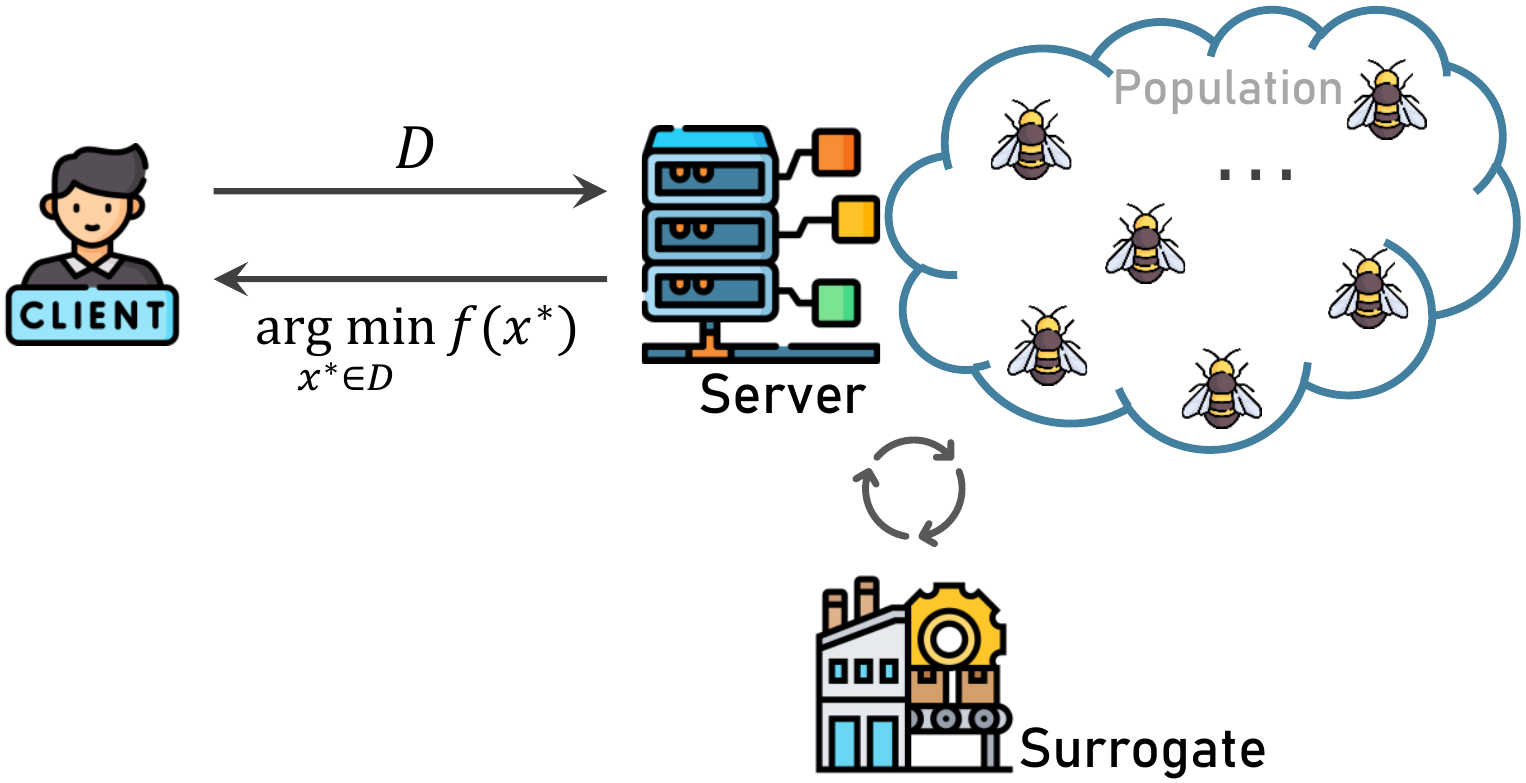}}
    \caption{Three typical optimization paradigms of evolutionary computation}
    \label{fig:opt_para}
\end{figure}
In this section, we formalize three typical optimization paradigms of evolutionary computation. Some notations are given as follows. We assume a client has an optimization problem $D$ to solve but he lacks programming skills and enough resources to solve the problem by himself. While $f$ is an objective function for the optimization problem. $x$ ($x \in D$) is a candidate solution for the optimization problem. $f(x)$ is called fitness or objective. $\oint_{\ell = 1} ^ {\ell = m}$ denotes $m$ rounds of iterative update of evolutionary algorithms. The optimal solution $x^*$ satisfies $\argmin_{x^* \in D} f(x^*)$. 

\subsection{Centralized Optimization Paradigm}
In the centralized optimization paradigm, the client outsources her optimization problem $D$ to centralized servers (e.g., cloud servers). The centralized server initializes a population consisting of $n$ individuals and makes each individual in the population maintain a candidate solution $x_i \in D$ ($i \in \{1, 2, \cdots, n\}$). The population jointly performs evolutionary operations to make their candidate solutions $\{x_1, x_2, \cdots, x_n\}$ approach the optimal solution $x^*$. Formally, the centralized optimization paradigm as shown in Fig. \ref{fig:opt_para}(a) can be formalized as
\begin{align}
    \argmin_{x^* \in D}f(x^*) \leftarrow \oint_{\ell = 1} ^ {\ell = m}\min\{f(x_1), f(x_2), \cdots, f(x_n)\}.
\end{align}

\subsection{Distributed Optimization Paradigm}
In the distributed optimization paradigm, the optimization problem $D$ is solved by a batch of distributed devices. Without loss of generality, assume $n$ distributed devices form a population. Each distributed device maintains a candidate solution $x_i$ ($i \in \{1, 2, \cdots, n\}$), while all distributed devices jointly perform evolutionary operations to approximate the optimal solution $x^*$. Formally, the distributed optimization paradigm as illustrated in Fig. \ref{fig:opt_para}(b) can be formalized as
\begin{align}
    \argmin_{x^* \in D}f(x^*) \leftarrow \oint_{\ell = 1} ^ {\ell = m}\min\{h_1(x_1), h_2(x_2), \cdots, h_n(x_n)\},
\end{align}
where $h_i(x_i)$ is the fitness of the $i$-th distributed device.

\subsection{Data-driven Optimization Paradigm}
The data-driven optimization paradigm tackles the issue that an optimization problem lacks objective functions, i.e., $f$ may not exist. In this case, historical data collected from simulations, physical experiments, or daily life is used to train a surrogate model. Then, the surrogate model is used instead of the missing objective function to evaluate the fitness of candidate solutions. In this paper, the surrogate model is denoted by $\mathcal{G}_{\theta}$, where $\theta$ denotes the model paradigms.

In the data-driven optimization paradigm, we consider three entities, i.e., a client with an optimization problem $D$, a server performing evolutionary algorithms, and a surrogate with the surrogate model. The server initializes a population comprising $n$ individuals and makes each individual in the population maintain a candidate solution $x_i \in D$ ($i \in \{1, 2, \cdots, n\}$). To evaluate the fitness of each candidate solution, the server cooperates with the surrogate to execute $\mathcal{G}_{\theta}(x_i)$ and obtains the fitness $y_i$, where $y_i \leftarrow \mathcal{G}_{\theta}(x_i)$. Formally, the data-driven optimization paradigm as depicted in Fig. \ref{fig:opt_para}(c) can be formalized as
\begin{align}
    \argmin_{x^* \in D}f(x^*) \leftarrow \oint_{\ell = 1} ^ {\ell = m}\min\{\mathcal{G}_{\theta}(x_1), \mathcal{G}_{\theta}(x_2), \cdots, \mathcal{G}_{\theta}(x_n)\},
\end{align}

\section{BOOM in Evolutionary Computation}
\begin{figure}[t]
    \centering
    \includegraphics[width=0.68\linewidth]{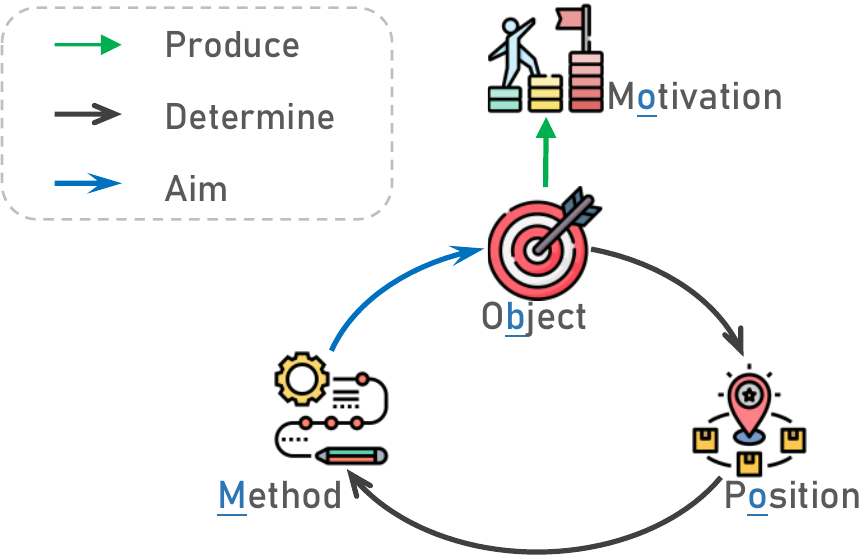}
    \caption{A general framework for privacy concerns in evolutionary computation}
    \label{fig:boom}
\end{figure}

This paper aims to bridge the gap between privacy protection and evolutionary computation and promotes the research on privacy-preserving evolutionary computation. To fully explore privacy concerns in evolutionary computation, we propose BOOM shown in Fig. \ref{fig:boom}, a general framework identifying privacy concerns in evolutionary computation, to cover all research issues of privacy protection in different evolutionary optimization paradigms. Specifically, BOOM defines the objective, the motivation, the position, and the method as follows.
\begin{itemize}
    \item Object: what data is regarded as private data.
    \item Motivation: why requires protecting private data.
    \item Position: where privacy-preserving technology is adopted.
    \item Method: what privacy-preserving technology makes a tradeoff between privacy protection and evolutionary computation.
\end{itemize}

The \textbf{object} confirms what data involved in an evolutionary computation procedure should be regarded as private data. In general, any privacy-preserving system first determines what data the system should protect. The object of privacy protection in evolutionary computation is just like the optimization object of an evolutionary algorithm. Only if the optimization object of an evolutionary algorithm is clear, the evolutionary algorithm knows how to approximate the optimal solution. Similarly, only if the object of privacy protection, privacy-preserving evolutionary computation knows how to protect private data.

Once the object of privacy protection is determined, there will produce a motivation for protecting private data. The \textbf{motivation} determines whether or not privacy-preserving evolutionary computation makes sense. In practice, it requires carefully highlighting the motivation that protects private data. When the object of private protection is clear, the most intuitive and biggest motivation is to prevent the leakage of private data.

The \textbf{position} states where privacy-preserving technology should be adopted. Evolutionary computation involves multiple types of optimization paradigms and multiple entities. Specifically, the position declares which entity in evolutionary computation should adopt privacy-preserving technology to protect private data and prevents other untrustworthy entities from obtaining private data. The position depends on the object of privacy protection. Different objects of privacy protection usually require different entities to adopt different privacy-preserving technologies. 

The \textbf{method} defines what privacy-preserving technologies to make a tradeoff between privacy protection and evolutionary computation. Different objects determine different positions, and then different positions determine different methods that should be adopted. Although there are many privacy-preserving technologies, there may be no existing privacy-preserving technologies to achieve privacy-preserving evolutionary computation directly. On the one hand, evolutionary computation usually involves multiple types of computations, such as addition, multiplication, division, comparison, and functional operation, which is difficult for existing privacy-preserving technologies to support all types of computations. On the other hand, privacy-preserving technologies not only require protecting private data but also do not lose the performance of evolutionary computation.

As shown in Fig. \ref{fig:boom}, the four items in BOOM are interrelated organic wholes. The object produces motivation. Also, the object determines the position, while the position determines the method. Finally, the method aims to ensure the object. In terms of research issues in privacy-preserving evolutionary computation, it is critical to clear the object of privacy protection. Apart from the object, the additional three items 
in BOOM can be regarded as three constraints that achieve the object.

\section{BOOM in Centralized Optimization Paradigm}
As illustrated in Fig. \ref{fig:opt_para}(a), a client specifies an optimization problem $D$. Then, the client asks a centralized server to perform evolutionary algorithms to tackle the optimization problem. Intuitively, in the centralized optimization paradigm, no privacy concern is involved as the centralized server only executes evolutionary operations and generates intermediate calculation data by itself. However, the centralized server takes as an input $D$ and outputs $\argmin_{x^* \in D}f(x^*)$. Particularly, the optimization problem consists of the specific problem (e.g., production planning, transportation planning and routing, neuroevolution) and its constraints, etc. Also, the output might be the optimal solution for the optimization problem, thus, it may be a neural network model, optimal planning of production, or optimal planning and routing of transportation. Correspondingly, BOOM issues in the centralized optimization paradigm include:

\textbf{Object}: This refers to what data intend to protect. In this case, it roughly comprises two types of data.
\begin{itemize}
    \item \textbf{Input}: Given a specific $D$, the centralized server always learns the detail of $D$, such as production planning, neuroevolution, transportation routing. Then, the centralized server can learn what constraints the customer uses to solve which problem. Thus, the input $D$ in the centralized optimization paradigm should be regarded as private data. In fact, for a privacy-preserving solution, the input is usually considered private data. For example, privacy-preserving training for machine learning \cite{mohassel2017secureml} always requires protecting the input data, i.e., training data. Also, privacy-preserving inference for machine learning \cite{uzun2021fuzzy} always needs to protect the input data, i.e., inference data.
    \item \textbf{Output}: The output of the centralized optimization paradigm is an approximate optimal solution. Obviously, no one wants to share her optimal production planning, optimal transportation routing, or optimal neural network. Therefore, the output $\argmin_{x^* \in D}f(x^*)$ in the centralized optimization paradigm should be regarded as private data. In practice, privacy-preserving training \cite{mohassel2017secureml} protects not only training data but also the output data, i.e., a trained machine learning model. In addition, privacy-preserving inference \cite{uzun2021fuzzy} protects not only inference data but also the inference result.
\end{itemize}

\textbf{Motivation}: This refers to why it requires protecting the input and output of the optimization problem. Roughly speaking, there are two motivations at least.
\begin{itemize}
    \item The centralized server is not always trusted. Thus, protecting the input and the output of the optimization problem is to prevent the centralized server from obtaining the detail of the optimization problem and the optimal solution.
    \item When the optimization problem is given to the centralized server, the client loses control of the optimization problem. Therefore, protecting the input and the output is to ensure the client can still control the optimization problem and its solution.
\end{itemize}

\textbf{Position}: In general, the centralized optimization paradigm comprises the client and the centralized server. According to the object of privacy protection including the input and the output, and the output generated by the centralized server through performing evolutionary operations, thus, the privacy-preserving technology should be adopted by the client and the centralized server. In other words, the position is two-aspect.
\begin{itemize}
    \item \textbf{Client side}: The client adopts the privacy-preserving technology to protect the input, i.e., the optimization problem $D$.
    \item \textbf{Server side}: The centralized server is enforced to adopt the privacy-preserving technology to protect the output, i.e., the approximate optimal solution $\argmin_{x^* \in D}f(x^*)$.
\end{itemize}

\textbf{Method}: This refers to a specific privacy-preserving technology that enables the centralized server to perform evolutionary computation effectively but protects the optimization problem and the approximate optimal solution.

\begin{figure}[t]
    \centering
    \includegraphics[width=0.89\linewidth]{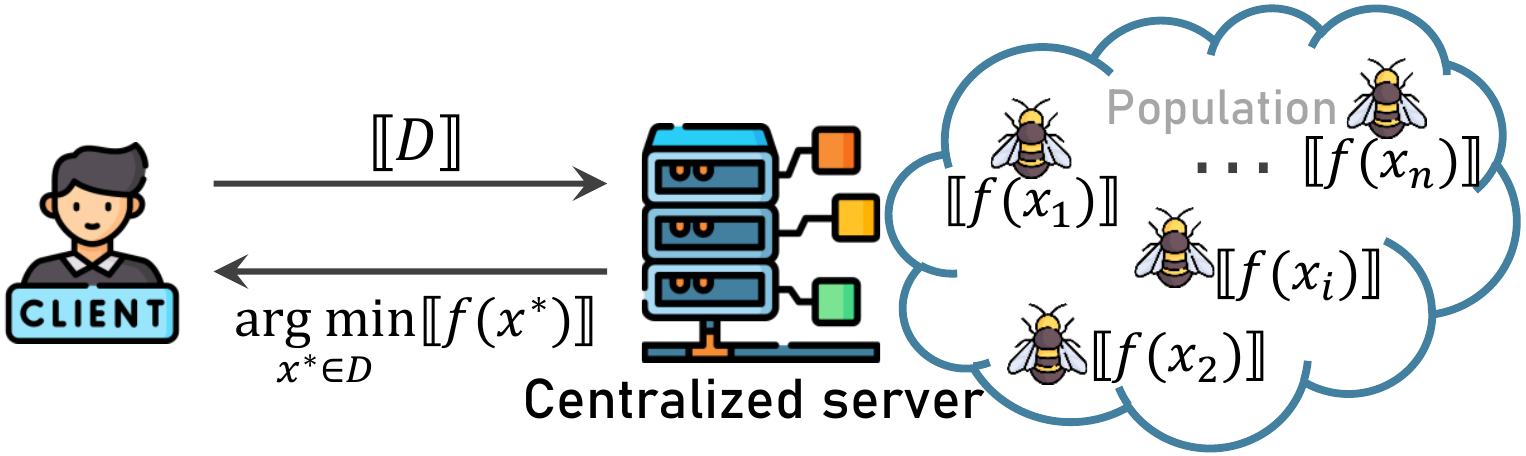}
    \caption{Privacy-preserving centralized optimization paradigm}
    \label{fig:ecop}
\end{figure}
Fig. \ref{fig:ecop} shows a possible privacy-preserving centralized optimization paradigm. Note that this paper uses $\lb \cdot \rb$ to denote a privacy-preserving technology. The privacy-preserving technology in the privacy-preserving centralized optimization paradigm needs to satisfy the following properties. On the one hand, the client can adopt privacy-preserving technology to protect $D$. Note that the centralized server needs to perform evolutionary operations based on $D$. Thus, the privacy-preserving technology protecting $D$ should support the centralized server to initialize a population and iteratively update the population without leaking $D$.

On the other hand, the purpose of evolutionary operations is essential to generate a better solution $x_i$ ($i \in \{1, \cdots, n\}$) that outputs less fitness. No matter what evolutionary algorithms, the key operation of approximating the optimal solution is to evaluate the fitness of the population. In other words, evolutionary algorithms always require executing comparison operations to evaluate fitness $f(x_i)$. Technically, the privacy-preserving technology enables the centralized server to maintain a population representing $n$ potential solutions $\{x_1, \cdots, x_n\}$ and to compute $\{\lb f(x_1) \rb, \cdots, \lb f(x_n) \rb\}$ effectively. Furthermore, the privacy-preserving technology supports the centralized server to compare $\{\lb f(x_1) \rb, \cdots, \lb f(x_n) \rb\}$ and output $\argmin_{x^* \in D}\lb f(x^*) \rb$.

In a word, the privacy-preserving technology (or say method) in the centralized optimization paradigm must support the following three types of computations.
\begin{itemize}
    \item \textbf{Encryption}: The method can encrypt the optimization problem effectively to protect private input.
    \item \textbf{Computation}: The method allows the centralized server to maintain a population presenting potential solutions without compromising the detail of $D$.
    \item \textbf{Comparison}: The method supports the comparison of the population's fitness but does not disclose any fitness to the centralized server. 
\end{itemize}

As far as the existing research is concerned, Jiang \textit{et al}. \cite{jiang2020privacy} concretized the privacy-preserving technology as somewhat homomorphic encryption, where evolutionary operations are outsourced to a cloud server. However, their proposed method fails to protect the optimization problem. Furthermore, their method cannot achieve a privacy-preserving selection operation as it does not support comparison over encrypted data. Also, Zhan \textit{et al}. \cite{zhan2021new} focused on the privacy protection of fitness and proposed a rank-based cryptographic function. However, the work \cite{zhan2021new} does not protect the optimization problem and does not give a detailed construct for the proposed cryptographic function. Although the work \cite{jiang2020privacy} and \cite{zhan2021new} explores privacy concerns in evolutionary computation, they do not support privacy-preserving evolutionary computation really. Recently, Zhao \textit{et al}. \cite{zhao2022evolution} concretized the privacy-preserving technology as a threshold Paillier cryptosystem. Particularly, $D$ is denoted by the traveling salesman problem. To protect $D$, the work \cite{zhao2022evolution} encrypts $D$ with the threshold Paillier cryptosystem. Moreover, the work designed a batch of privacy-preserving computation protocols based on the threshold Paillier cryptosystem on a twin-cloud server architecture to achieve evolutionary operations over encrypted data and output an encrypted solution.

\section{BOOM in Distributed Optimization Paradigm}
As illustrated in Fig. \ref{fig:opt_para}(b), a client specifies an optimization problem $D$. Then, the client asks distributed devices to jointly perform an evolutionary algorithm to tackle the optimization problem. Different from the centralized optimization paradigm, the distributed optimization paradigm takes as input $D$ and each distributed device's solution, where the latter is iteratively updated based on each distributed device's local data (e.g., a locally optimal solution) and a global optimization solution. Thus, the distributed optimization paradigm suffers from more complex privacy concerns than the centralized optimization paradigm as it involves more parties. Correspondingly, BOOM issues in the centralized optimization paradigm include:

\textbf{Object}: This refers to what data intend to protect. In this case, it roughly comprises three types of data.
\begin{itemize}
    \item \textbf{Input}: Given a specific $D$, distributed devices always learn the detail of $D$, such as production planning, neuroevolution, transportation routing. Then, distributed devices can learn what constraints the customer uses to solve which problem. Thus, the input $D$ in the distributed optimization paradigm should be regarded as private data.
    \item \textbf{Output}: The output of the distributed optimization paradigm is an approximate optimal solution. Obviously, no one wants to share her optimal production planning, optimal transportation routing, or optimal neural network. Therefore, the output $\argmin_{x^* \in D}f(x^*)$ in the centralized optimization paradigm should be regarded as private data.
    \item \textbf{Local solution}: The local solution means the potentially optimal solution that is maintained by each distributed device. If the approximate optimal solution is private data, the local solution should be also private data. Furthermore, each local solution is generated based on the distributed device's local data and the global optimization solution. Each distributed device's local data is its private data. Thus, each distributed device's local solution is regarded as private data.
\end{itemize}

\textbf{Motivation}: This refers to why it requires protecting the input and output of the optimization problem as well as the local solution. Roughly speaking, there are two motivations at least.
\begin{itemize}
    \item The distributed device is not always trusted. On the one hand, protecting the input and the output of the optimization problem is to prevent untrustworthy distributed devices from obtaining the detail of the optimization problem and the optimal solution. On the other hand, protecting the local solution is to prevent untrustworthy distributed devices from obtaining other devices' solutions, and to avoid untrustworthy distributed devices hitchhiking.
    \item When the optimization problem is given to distributed devices, the client loses control of the optimization problem. Therefore, protecting the input and the output is to ensure the client can still control the optimization problem and its solution.
\end{itemize}

\textbf{Position}: According to the object of privacy protection including the input, the output, and the local solution, thus, the privacy-preserving technology should be adopted by the client and each distributed device. In other words, the position is two-aspect.
\begin{itemize}
    \item \textbf{Client side}: The client adopts the privacy-preserving technology to protect the input, i.e., the optimization problem $D$.
    \item \textbf{Distributed device side}: Each distributed device adopts the privacy-preserving technology to protect its local solution and the output, i.e., the approximate optimal solution $\argmin_{x^* \in D}f(x^*)$.
\end{itemize}

\textbf{Method}: This refers to a specific privacy-preserving technology that enables the centralized server to perform evolutionary computation effectively but protects the object.

\begin{figure}[t]
    \centering
    \includegraphics[width=0.83\linewidth]{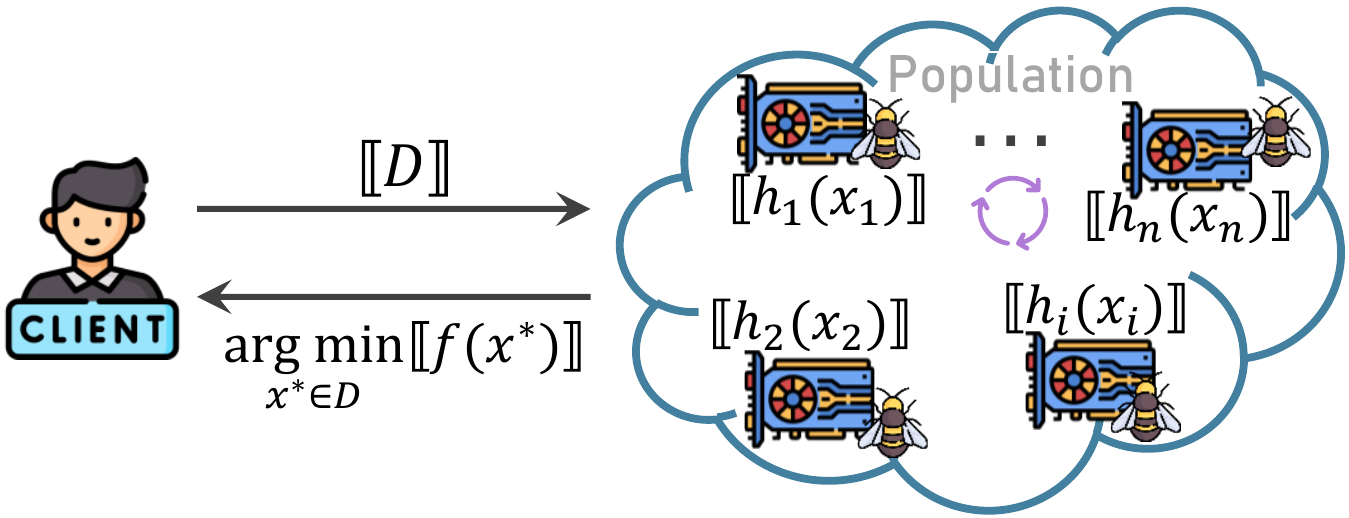}
    \caption{Privacy-preserving distributed optimization paradigm}
    \label{fig:edop}
\end{figure}
Fig. \ref{fig:edop} illustrates a possible privacy-preserving distributed optimization paradigm. The privacy-preserving technology in the privacy-preserving distributed optimization paradigm needs to satisfy the following properties. On the one hand, the client can adopt the privacy-preserving technology to protect $D$. As each distributed device needs to perform evolutionary operations based on $D$, the privacy-preserving technology protecting $D$ should allow the distributed devices to initialize a local solution and iteratively update the local solution without leaking $D$.

On the other hand, the privacy-preserving technology enables multiple distributed devices to evaluate their fitness effectively but does not compromise any distributed device's local solution.

In a word, the privacy-preserving technology (or say method) in the distributed optimization paradigm must support the following three types of computations.
\begin{itemize}
    \item \textbf{Encryption}: The method can encrypt the optimization problem effectively to protect private input.
    \item \textbf{Computation}: The method allows multiple distributed devices to maintain local solutions without leaking $D$.
    \item \textbf{Comparison}: The method supports the comparison of the fitness but does not disclose any distributed device's local solution. 
\end{itemize}

As far as the existing research is concerned, Han \textit{et al}. \cite{han2007privacy} considered two parties jointly performing a privacy-preserving evolutionary algorithm to approximate the optimal solution, where each party's fitness is securely evaluated without each party's local solution. After that, Funke \textit{et al}. \cite{funke2010privacy} concretized the privacy-preserving technology as Yao’s protocol, additive secret sharing, and the Paillier cryptosystem and proposed a privacy-preserving two-party multi-object evolutionary algorithm. Whether \cite{han2007privacy} or \cite{funke2010privacy} assumes two parties share the same optimization problem and fails to consider untrustworthy parties. Recently, Zhao \textit{et al}. \cite{zhao2022primpso} formulated the privacy-preserving technology as the Paillier cryptosystem and secure multi-party computation and designed a distributed particle swarm optimization algorithm.

\section{BOOM in Data-Driven Optimization Paradigm}
As depicted in Fig. \ref{fig:opt_para}(c), a client specifies an optimization problem $D$, where $D$ lacks an object function evaluating fitness, and the fitness evaluation of the optimization problem relies on historical data. Then, the client asks a server (e.g., a cloud server) to tackle the optimization problem, while the server is assisted by a surrogate with a surrogate model evaluating fitness. The server takes $D$ as an input and outputs $\argmin_{x^* \in D}f(x^*)$. Particularly, the optimization problem consists of the specific problem (e.g., production planning, transportation planning and routing, neuroevolution) and its constraints, etc. Also, the output might be the optimal solution for the optimization problem, thus, it may be a neural network model, optimal planning of production, or optimal planning and routing of transportation. Correspondingly, BOOM issues in the data-driven optimization paradigm include:

\textbf{Object}: This refers to what data intend to protect. In this case, it roughly comprises four types of data.
\begin{itemize}
    \item \textbf{Input}: Given a specific $D$, the server always learns the detail of $D$, such as production planning, neuroevolution, transportation routing. Then, the server can learn what constraints the customer uses to solve which problem. Thus, the input $D$ in the data-driven optimization paradigm should be regarded as private data.
    \item \textbf{Output}: The output of the data-driven optimization paradigm is an approximate optimal solution. Obviously, no one wants to share her optimal production planning, optimal transportation routing, or optimal neural network. Therefore, the output $\argmin_{x^* \in D}f(x^*)$ in the data-driven optimization paradigm should be regarded as private data.
    \item \textbf{Potential solution}: The server initializes a population representing potential solutions and iteratively updates the population, where the surrogate takes the potential solution as input and performs the evolutionary operation of fitness evaluation. If the approximate optimal solution is private data, the potential solution should be also private data. The surrogate evaluates the fitness of the potential solution, which is essentially an inference operation based on machine learning. The privacy-preserving inference of machine learning \cite{uzun2021fuzzy} always requires protecting the input data. Thus, potential solutions maintained by the server are regarded as private data.
    \item \textbf{Surrogate model}: The surrogate model is a machine learning model. The machine learning model learns knowledge from abundant training data and transforms the knowledge into model parameters. Furthermore, training a machine learning model consumes a mass of resources. Hence, in the research field of privacy-preserving machine learning \cite{mohassel2017secureml,uzun2021fuzzy}, the machine learning model is always regarded as private data.
\end{itemize}

\textbf{Motivation}: This refers to why it requires protecting the input and output of the optimization problem as well as the local solution. There are three motivations at least.
\begin{itemize}
    \item The server is not always trusted. Thus, protecting the input and the output of the optimization problem is to prevent an untrustworthy server from obtaining the detail of the optimization problem and the optimal solution.
    \item When the optimization problem is given to the server, the client loses control of the optimization problem. Therefore, protecting the input and the output is to ensure the client can still control the optimization problem and its solution.
    \item The surrogate might be untrustworthy. Thus, protecting the potential solution is to avoid disclosing inference data, i.e., the potential solution.
\end{itemize}

\textbf{Position}: According to the object of privacy protection including the input, the output, and the potential solution, thus, the privacy-preserving technology should be adopted by the client, the server, and the surrogate. Specifically, the position is three-aspect.
\begin{itemize}
    \item \textbf{Client side}: The client employs the privacy-preserving technology to protect the input, i.e., the optimization problem $D$.
    \item \textbf{Server side}: The server adopts the privacy-preserving technology to protect its potential solutions and the output, i.e., the approximate optimal solution $\argmin_{x^* \in D}f(x^*)$.
    \item \textbf{Surrogate side}: The surrogate utilizes privacy-preserving technology to protect its surrogate model and inference results.
\end{itemize}

\textbf{Method}: This refers to a specific privacy-preserving technology that enables the server and the surrogate to jointly perform evolutionary computation effectively but protects the object.

\begin{figure}[t]
    \centering
    \includegraphics[width=0.83\linewidth]{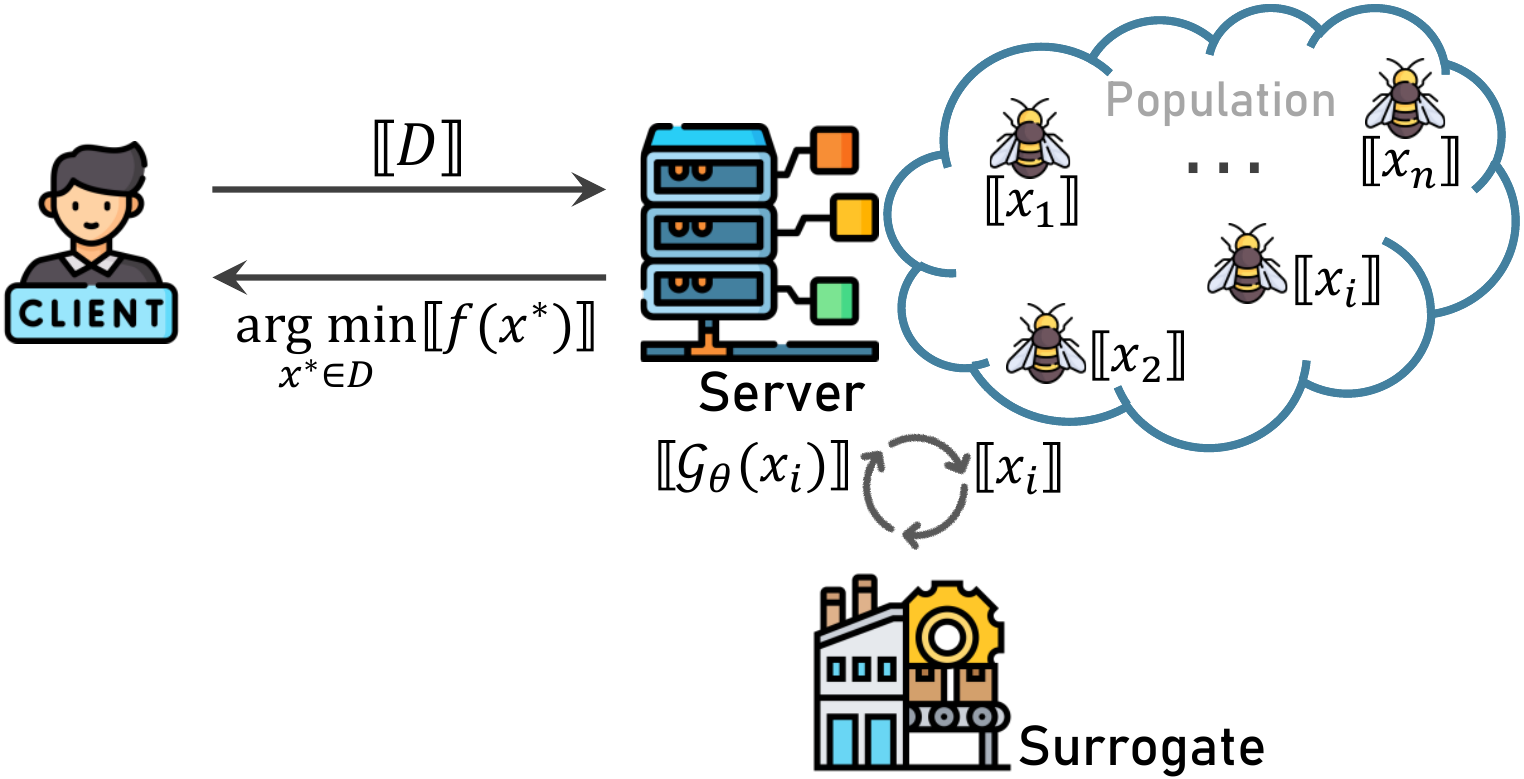}
    \caption{Privacy-preserving data-driven optimization paradigm}
    \label{fig:eddop}
\end{figure}
Fig. \ref{fig:eddop} shows a possible privacy-preserving data-driven optimization paradigm. The privacy-preserving technology in the privacy-preserving data-driven optimization paradigm needs to satisfy the following properties. On the one hand, the client can adopt privacy-preserving technology to protect $D$. As the server needs to perform evolutionary operations based on $D$, the privacy-preserving technology protecting $D$ should allow the server to initialize a population and iteratively update the population without leaking $D$.

On the other hand, the privacy-preserving technology enables the server to protect its potential solutions (or say inference data), it also protects the surrogate model and inference results.

In short, the privacy-preserving technology (or say method) in the data-driven optimization paradigm must support the following three types of computations.
\begin{itemize}
    \item \textbf{Encryption}: The method can encrypt the optimization problem effectively to protect private input.
    \item \textbf{Computation}: The method enables the server to maintain the population without compromising $D$.
    \item \textbf{Comparison}: The method achieves the comparison of the fitness but does not disclose any fitness to the server.
    \item \textbf{Inference}: The method supports the surrogate to perform privacy-preserving inference of machine learning.
\end{itemize}

As far as the existing research is concerned, to the best of our knowledge, there is no existing solution that exploits privacy-preserving data-driven optimization. One possible explanation is that privacy concerns in data-driven evolutionary computation involve not only privacy concerns of evolutionary computation but also privacy concerns of machine learning. In other words, data-driven evolutionary computation suffers from more and more complex privacy concerns.

\section{Future Research Directions}
With BOOM issues elaborated on in previous sections, it can be seen that privacy concerns in evolutionary computation do not be fully exploited. We argue that research issues of privacy protection in evolutionary computation are still not well exploited as the object and the motivation of privacy protection are unclear, the position for adopting privacy-preserving technology is varying, and the effectiveness and efficient method of achieving privacy protection is nonexistent. Based on the identified gaps and typical optimization paradigms of evolutionary computation, we foresee the following research directions for privacy-preserving evolutionary computation.
\begin{itemize}
    \item \textbf{Evolution as a Service}: Currently, almost all evolutionary algorithms lose sight of a user lacking enough capability to implement evolutionary algorithms but she requires to solve an optimization problem through evolutionary algorithms. Computation outsourcing is an emerging computing paradigm to tackle the problem that the user lacks enough capability or resources to perform computations. Thus, a cloud server with sufficient computing power and resources can provide an evolutionary computation service for the user. To prevent an untrustworthy cloud server, secure computation outsourcing has received wide attention. The combination of evolutionary computation and secure computation outsourcing is likely to achieve a tradeoff between the performance of the centralized optimization paradigm and its privacy protection. Particularly, the computing paradigm that a cloud server provides privacy-preserving evolutionary computation servers can be called evolution as a service.
    \item \textbf{Privacy-preserving Federated Optimization}: In essence, the distributed optimization paradigm is that distributed devices jointly solve the same optimization problem in a federated manner. Federated learning, a collaborative learning manner, makes distributed participants jointly train the same machine learning model \cite{bonawitz2019towards}. Federated learning can protect each participant's training data and the machine learning model. Inspired by federated learning, privacy-preserving federated optimization is likely to provide a solution for making a tradeoff between the performance of the distributed optimization paradigm and its privacy protection. Specifically, privacy-preserving federated optimization allows each distributed device to generate its local solution based on its historical data and the global solution. To protect privacy, each distributed device never shares its local historical data with others, while each distributed device cannot learn the global solution. Inspired by existing privacy-preserving technology in federated learning, homomorphic encryption, secure multi-party computation, and differential privacy are potentially underlying technologies to achieve privacy-preserving federated optimization, where the key point of the privacy-preserving technology is to support multiple computations without compromising privacy.
    \item \textbf{Privacy-preserving Data-driven Optimization}: Currently, data-driven optimization relies on a surrogate to train a surrogate model based on historical data collected from simulations, physical experiments, etc. To achieve a tradeoff between the performance of the data-driven optimization paradigm and its privacy, privacy-preserving data-driven optimization is likely to be a potential solution. Specifically, the privacy-preserving data-driven optimization might exploit a uniform framework that balances privacy-preserving evolutionary computation and privacy-preserving inference. In terms of technology, privacy-preserving technology (e.g., the combination of secret sharing, homomorphic encryption, and garbled circuit) supporting privacy-preserving inference and privacy-preserving evolutionary operations is a potential method.
\end{itemize}

\section{Conclusion}
With privacy issues in all kinds of computation fields (e.g., machine learning, crowdsourcing, search) having been widely concerned, privacy-preserving technology plus the type of computation has become a popular research topic. However, privacy concerns in evolutionary computation fail to be fully explored, but it can be expected that the scope and depth of the research on privacy-preserving evolutionary computation will further expand in the years to come. To sort out the research issues in this emerging research domain, this paper proposes BOOM (i.e., object, motivation, position, and method), and adopts BOOM to characterize privacy concerns in three typical optimization paradigms of evolutionary computation. By introducing BOOM, most of the research issues in privacy-preserving evolutionary computation are revealed in a structured manner, and several research directions are identified. this paper aims to provide guidelines and insights for interesting researchers in the emerging privacy-preserving evolutionary computation field.

\bibliographystyle{IEEEtran}
\bibliography{IEEEabrv,ref}

\end{document}